\def\year{2020}\relax
\documentclass[letterpaper]{article} 
\usepackage{aaai20}  
\usepackage{times}  
\usepackage{helvet} 
\usepackage{courier}  
\usepackage[hyphens]{url}  
\usepackage{graphicx} 
\usepackage{graphicx}  
\usepackage{slashbox}
\usepackage{mathtools}
\usepackage{amsmath, amsthm, amssymb, amsfonts, amsopn}
\usepackage{graphicx} 
\usepackage{textcomp}
\usepackage{xfrac}
\usepackage{bbm}
\usepackage{overpic}
\usepackage{subfig}
\usepackage{multirow}
\newcommand{\supl}[1]{{#1}}
\newcommand{\hlr}[1]{{\color{red}\textbf{#1}}}
\newcommand{\hlg}[1]{{\color{green}\textbf{#1}}}

\title{Learning Part Generation and Assembly for Structure-aware Shape Synthesis}
\author{Jun Li, Chengjie Niu, Kai Xu\thanks{Corresponding author: Kai Xu (kevin.kai.xu@gmail.com)}\\
National University of Defense Technology}

\begin{document}

\maketitle


\begin{abstract}\vspace{-3pt}
Learning powerful deep generative models for 3D shape synthesis
is largely hindered by the difficulty in ensuring plausibility
encompassing correct topology and reasonable geometry. Indeed, learning the distribution of plausible 3D shapes seems a daunting task for the holistic approaches, given the significant topological variations of 3D objects even within the same category. Enlightened by the fact that 3D shape structure is characterized as part composition and placement, we propose to model 3D shape variations with a \emph{part-aware} deep generative network, coined as \emph{PAGENet}. The network is composed of an array of per-part VAE-GANs, generating semantic parts composing a complete shape, followed by a part assembly module that estimates a transformation for each part to correlate and assemble them into a plausible structure. Through delegating the learning of part composition and part placement into separate networks, the difficulty of modeling structural variations of 3D shapes is greatly reduced. We demonstrate through both qualitative and quantitative evaluations that \emph{PAGENet} generates 3D shapes with plausible, diverse and detailed structure, and show two applications, i.e., semantic shape segmentation and part-based shape editing.\vspace{-3pt}
\end{abstract}


\vspace{-8pt}
\section{Introduction}
\begin{figure}[t]
	\begin{center}
		\includegraphics[width=1.0\linewidth]{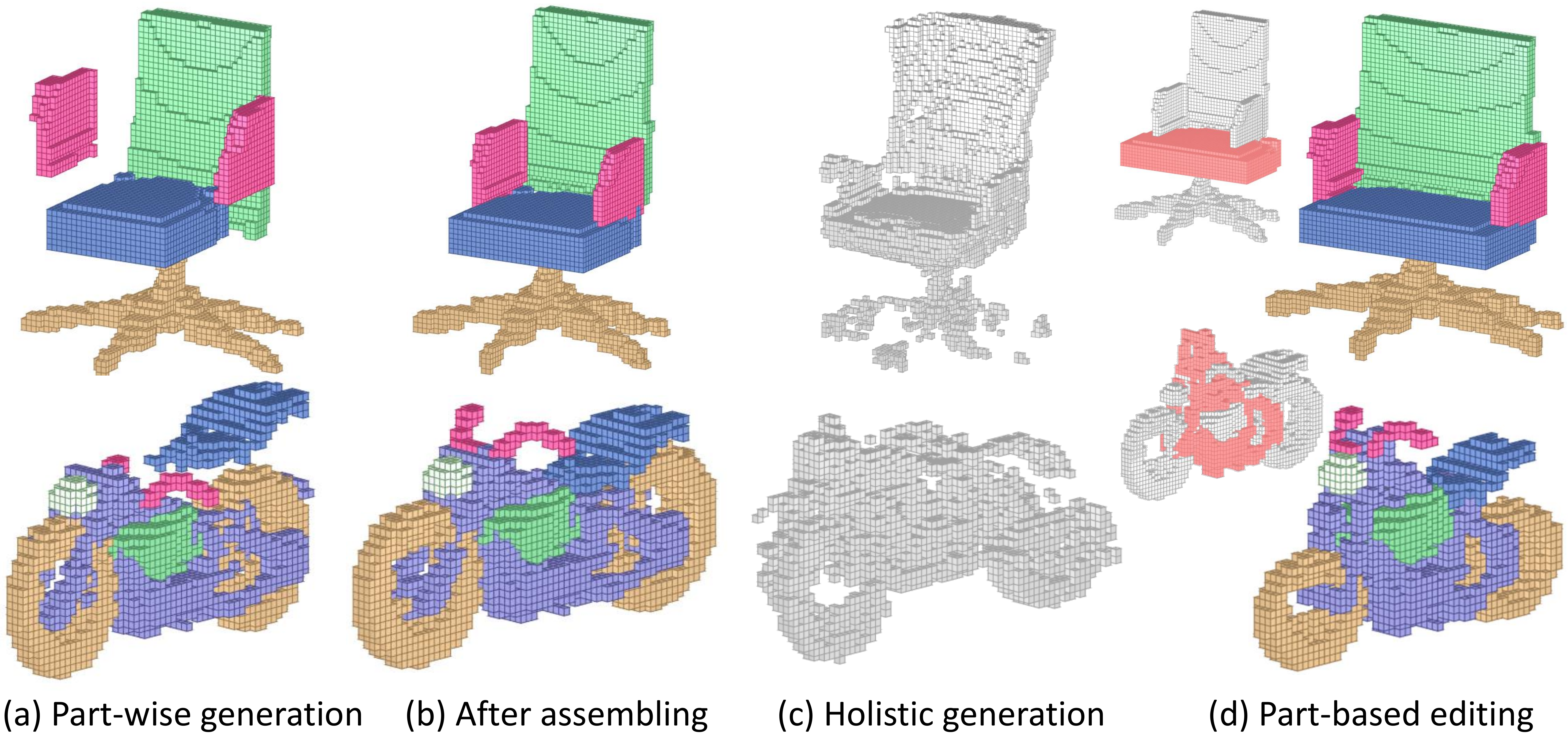}
	\end{center}\vspace{-12pt}
	\caption{PAGENet first generates the semantic parts (a) and then assembles them with proper transformations, forming a structurally valid 3D shape (b) in contrast to the results by holistic generation (c). The learned model supports part-based shape editing (d): The anchor part (the pink parts in the inset) is edited by the user and all parts are automatically deformed and re-assembled into a plausible shape.}
	\label{fig:teaser}\vspace{-16pt}
\end{figure}

Learning deep generative models for 3D shape synthesis has been attracting increasing research interest lately.
Despite the notable progress made since the seminal work of~\cite{wu2016learning},
one major challenge still demanding more attention is
how to ensure the \emph{structural correctness} of the generated shapes.
Most existing generative models are structure-oblivious. They tend to generate
3D shapes in a holistic manner, without comprehending its compositional parts explicitly.
Consequently, when generating 3D shapes with detailed structures,
the details of complicated part structures are often blurred or even messed up (see Figure~\ref{fig:teaser}(c)).

To alleviate this issue, one common practice is to increase the shape resolution
to better capture fine-grained details, with the cost of increased training examples and learning time.
The main reason behind is that the 3D shape representation employed by those models,
e.g., volume or point cloud, are oblivious to shape structures.
With such representations, part information is not encoded and thus cannot be decoded either during the generation process.

In the field 3D shape analysis, a consensus about the ``structure'' of a shape is the combination
of part composition and the relative positioning between parts~\cite{mitra2013structure,xu2016data}.
Following that insight, we approach the modeling of structural variations of 3D shapes through learning a
\emph{part-aware} generative network, named as PAGENet.
The model is thus aware of what parts it is generating, through
a semantic-part-wise generation process. On the other hand, the model should be able to preserve the relative spatial placement between the generated parts, according to some learned part placement prior.
Such placement determines how adjacency semantic parts assemble together in forming a structurally valid shape.

PAGENet is composed of an array of \emph{part generators}, each of which is
a combination of variational auto-encoder (VAE) and generative adversarial network (GAN) trained for generating a specified semantic part of the target shape category,
followed by a \emph{part assembler} that estimates a transformation for each part used to assemble
them into a valid shape structure (Figure~\ref{fig:teaser}(a,b)). In this work, PAGENet is realized in the volumetric setting; it can be easily extended to other representations such as point cloud (see Figure~\ref{fig:hirespc}).

Our model delegates the generation of parts and relations on separate networks,
greatly reducing the modeling difficulty of 3D shape structures.
3D structure variation is modeled by the concatenation of the latent vectors
of all part generators, forming a \emph{structured latent
space} of 3D shapes with different dimensions of the latent vector
controlling the generation of different parts.
This facilitates part-based shape editing and latent-space part reshuffling of the generated shapes with structure preservation. Figure~\ref{fig:teaser}(d) shows two examples of part-based editing: Based on our part assembler, edit over anchor parts can be propagated to other parts, resulting in structurally reasonable edited shapes.

Our main contributions include:
\begin{itemize}
  \item A divide-and-conquer approach to learning a deep generative model of structure-aware shape generation.
  \item A part assembly module learned to correlate the generated semantic parts with their assembling transformations.
  \item Two applications including semantic shape segmentation and part-based structure-aware shape editing.
\end{itemize}


\vspace{-6pt}
\section{Related work}

\paragraph{Deep generative models of 3D shapes.}
Deep generative models for 3D shape generation have been developed based on various 3D representations, such as volumetric grids~\cite{wu2016learning,girdhar2016learning,Wang-2017-OCNN,riegler2017octnet}, point clouds~\cite{fan2016point,Achlioptas2018}, surface meshes~\cite{groueix2018papier,wang2018pixel2mesh}, implicit functions~\cite{chen2018learning,park2019deepsdf}, and multi-view images~\cite{ArsalanCVPR2017}.
Common to these works is that shape variability is modeled in a holistic, structure-oblivious fashion,
which is mainly due to the limited options of deep-learning-friendly 3D shape representations.

\vspace{-5pt}
\paragraph{Structure-aware 3D shape synthesis.}
Research on \emph{deep generative models} for structure-aware shape synthesis
starts to gain increasing attention recently.
Huang et al.~\shortcite{Huang:2015:deeplearningsurfaces} propose a deep generative model
based on part-based templates learned \emph{a priori}, which is not end-to-end trainable.
Li et al.~\shortcite{li2017grass} propose the first deep generative model of 3D shape structures.
They employ recursive neural network to achieve hierarchical encoding and decoding of parts and relations.
The binary-tree-based architecture is extended to N-ary in~\cite{mo2019structurenet}.
However, these models do not explicitly ensure a quality part assembly as in our method; see Section~\ref{sec:result} for a comparison.
In \cite{zou20173d}, a sequential part generation model is learned with recurrent neural networks, which produces parts as cuboids only.

Nash and Williams~\shortcite{Nash2017} propose ShapeVAE to generate part-segmented 3D objects. Their model is trained using shapes with dense point correspondence. On the contrary, our model requires only part level correspondence.
Wu et al.~\shortcite{Wu2018Struct} couples the synthesis of intra-part geometry
and inter-part structure. Similar idea is proposed in~\cite{Balashova2018} where
landmark-based structure priors are used for structure-aware shape generation.
In \cite{G2L18}, 3D shapes are generated with part labeling based on GAN and then refined using a pre-trained part refiner.
Our method takes a reverse process where we first generate parts and then their assembling transformations.

More closely related are the two concurrent works of part-based shape synthesis in~\cite{Schor2019} and~\cite{Dubrovina2019}.
Schor et al.~\shortcite{Schor2019} train part-wise generators and a part composition network
for the generation of 3D point clouds.
Dubrovina et al.~\shortcite{Dubrovina2019} propose a nice decomposer-composer network to learn a factorized shape embedding space for part-based 3D shape modeling. Different from our work, however, their model is not a generative one.
In their work, novel shapes are synthesized through randomly sampling and assembling the pre-exiting parts embedded in the factorized latent space.


\begin{figure}[t]
\begin{center}
\includegraphics[width=0.9\linewidth]{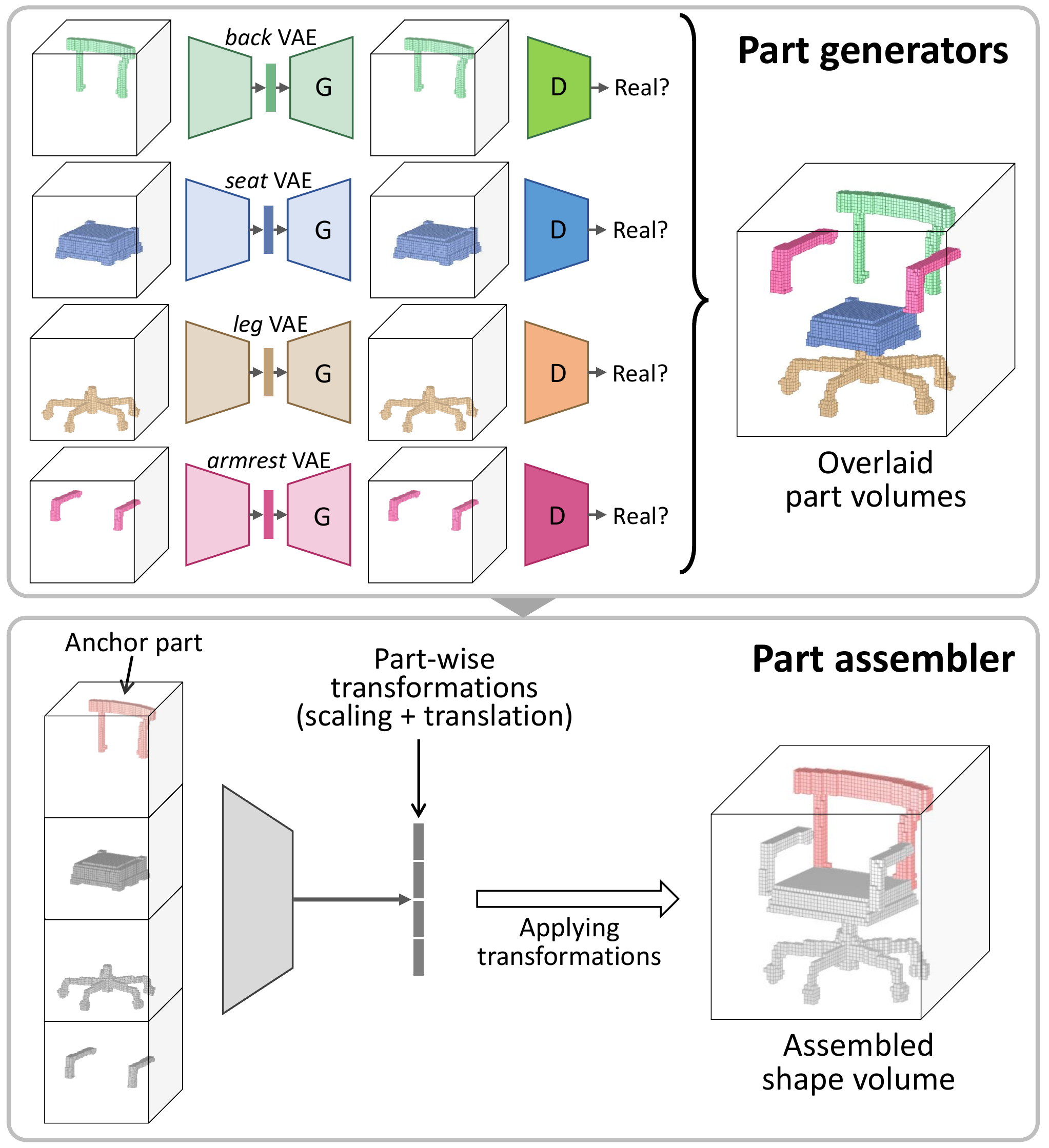}
\end{center}\vspace{-12pt}
   \caption{Top: Training the part generators (one VAE-GAN per semantic part). Since parts are generated independently, there could be mismatching and disconnection (see the overlaid volumes). Bottom: A part assembler is trained to regress a transformation (scaling + translation) for each part, which assembly the parts into a plausible shape.}
\label{fig:overview}\vspace{-10pt}
\end{figure}

\section{Method}
\subsection{Network architecture}
Our network architecture (Figure~\ref{fig:overview}) is composed of two modules: a part-wise generative network and a part assembly module. The part-wise generative network contains $K$ part generators, each for one of the $K$ predefined semantic part labels (e.g., back, seat, leg and armrest for a chair). Each part generator is trained to generate a volume of a specific part from a random vector.
Taking the generated volumes for all the parts as input, the part assembler predicts a transformation (scaling + translation) for each part, to assemble the parts into a complete shape with proper part scaling and inter-part connection.

\subsection{Part generators}
For each semantic part, we train a generative network of 3D volumes, which is a combination of variational auto-encoder (VAE) and generative adversarial network (GAN), or VAE-GAN. The VAE part comprises of an encoder and a decoder of 3D volumes with a resolution of $64\times64\times64$. The dimension of the latent vector is $50$. Similar to~\cite{wu2016learning}, the encoder consists of five volumetric fully
convolutional layers with a kernel size of $4\times4\times4$ and a stride of $2$.
Batch normalization and ReLU layers are inserted between convolutional layers.
The decoder / generator simply reverses the encoder, except that a $Sigmoid$ nonlinearity is used in the last layer.
Following the decoder, a discriminator is devised by reusing the encoder architecture (with an extra $1$D MLP output layer). The discriminator is trained to tell whether a given part volume is real, i.e., the voxelization of a real shape part, or fake, i.e., a volume generated by the generator.

Therefore, the loss function for a part generator consists of three terms: a part volume reconstruction loss $L_{\text{recon}}$, a Kullback-Leibler divergence loss $L_{\text{KL}}$ and a adversarial loss $L_{\text{adv}}$.
In addition, we introduce a reflective symmetry loss $L_{\text{ref}}$ to penalize generating asymmetric parts. This loss would help regularize the part generation, since most parts are reflective symmetric. For those asymmetric parts, the weight of $L_{\text{ref}}$ is set to $0$.
In summary, the loss is:
\begin{equation}
L_{\text{partgen}}(p) = L_{\text{recon}}+\alpha_1L_{\text{KL}}+\alpha_2L_{\text{adv}}+\delta_{\text{ref}}(p)L_{\text{ref}},
\label{eq:vae_loss}
\end{equation}
where $L_{\text{recon}} = D(x_\text{in},x_\text{out})$ measures the mean-square-error (MSE) loss between the input volume $x_\text{in}$ and output volume $x_\text{out}$; $L_\text{ref}(x) = D(x,ref(x))$ is the MSE loss between a volume $x$ and its reflection $ref(x)$ about the reflection plane of the input shape; $\delta_{\text{ref}}(p)$ is a Kronecker delta function indicating whether part $p$ shares reflective symmetry with the full shape.
In training, we detect reflective symmetry using the method in~\cite{mitra2006partial} for a training shape and its semantic parts, to evaluate $\delta_{\text{ref}}$ for each part.

For the adversarial training, we follow WGAN-GP~\cite{GulrajaniNIPS17} which improves Wasserstein GAN~\cite{arjovsky2017wasserstein} with a gradient penalty to train our generative model,
\begin{equation}
L_{\text{adv}} = \mathop{E[D(\widetilde{x})]}\limits_{\widetilde{x}\sim P_g} - \mathop{E[D(x)]}\limits_{x \sim P_r} + \lambda \mathop{E}\limits_{\hat{x}}[(\left \| \nabla_{\hat{x}}D(\hat{x})\right \|_2-1)^2],
\label{eq:gan_loss}
\end{equation}
where $D$ is the discriminator, $P_g$ and $P_r$ are the distributions of generated part volumes and real part volumes, respectively. The last term is the gradient penalty and $\hat{x}$ is sampled uniformly along straight lines between pairs of points from the data distribution $P_r$ and the generator distribution $P_g$. The discriminator attempts to minimize $L_\text{adv}$ while the generator maximizes the first term in Equation (\ref{eq:gan_loss}).

\subsection{Part assembler}
Since the part volumes are generated independently, their scales may not match with each other and their positions may disconnect adjacent parts. Taking $K$ part volumes generated from the part-wise generative network, part assembler regresses a transformation, including a scaling and a translation, for each part. It relates different semantic parts, with proper resizing and repositioning, to assemble them into a valid and complete shape volume. Essentially, it learns the spatial relations between semantic parts (or part arrangements) in terms of relative size and position, as well as mutual connection between different parts.

The part assembler takes $K$ part volumes as input, which amount to a $64\times64\times64\times K$ input tensor. The input tensor is passed through five volumetric fully convolutional layers of kernel sizes $4\times4\times4$ with a stride of $2$. Similar to part encoders, batch normalization and ReLU layers are used between convolutional layers. In the last layer, a sigmoid layer is added to regress the scaling and translation parameters. To ease the training, we normalize all scaling and translation parameters into $[0,1]$, based on the allowed range of scaling ($[0.5,1.5]$) and translation ($[-20,20]$, with the unit being voxel size). The actual values of scaling and translation parameters are recovered when being applied.

\begin{figure}[t]
	\begin{center}
		\includegraphics[width=1.0\linewidth]{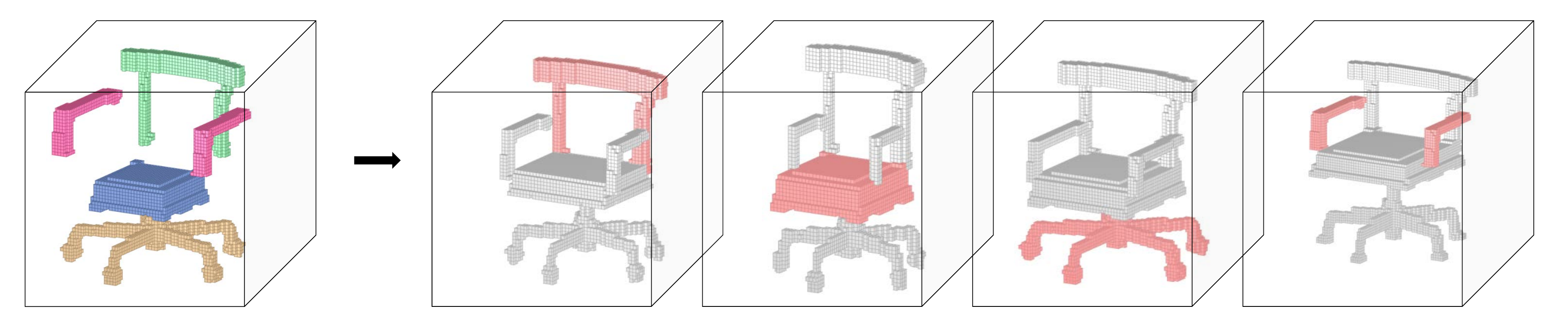}
	\end{center}\vspace{-12pt}
	\caption{Reasonable assembly result can be produced with different parts serving as anchor (highlighted in red color).}
	\label{fig:assembly_example}\vspace{-8pt}
\end{figure}

\begin{figure}[t]
\begin{center}
\includegraphics[width=0.95\linewidth]{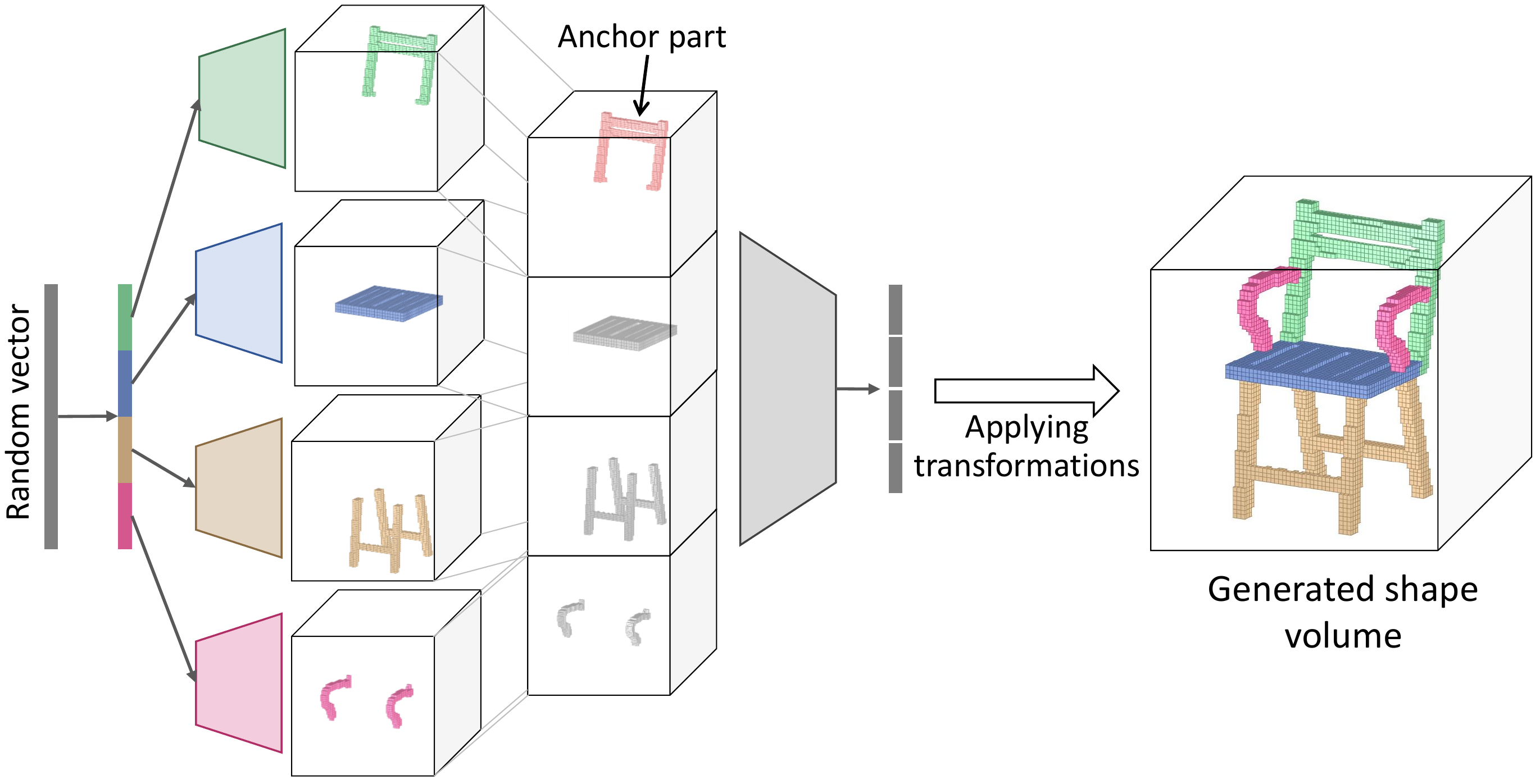}
\end{center}\vspace{-12pt}
   \caption{Part-aware 3D shape generation. Given a random vector, our network splits it into several sections and pass them to respective part decoders, yielding a set of semantic parts. The parts are then assembled together based on the predicted per-part transformations.}
\label{fig:test}\vspace{-10pt}
\end{figure}

\vspace{-8pt}
\paragraph{Anchored transformation.}
Given $K$ part volumes, the transformations assembling them together is not unique. Taking the chair model in Figure~\ref{fig:assembly_example} as an example, the chair seat can be stretched to match the back, while the back can also be shrunk to conform to the seat; both result in a valid shape structure. This also adds some diversity to the generation.
To make the transformation estimation determined and the assembly network easier to train, we introduce an extra input to the part assembler to indicate a \emph{anchor part}. When estimating part transformations, the anchor part is kept fixed (with an identity transformation) while all the other parts are transformed to match the anchor.
To do this, one option is to input an indicator vector (a one-hot vector with the corresponding part being $1$).
However, the dimension of this indicator vector is too small, making its information easily overwhelmed by the large tensor of part volumes. Therefore, we opt to infuse anchor information by setting the occupied voxels in the anchor part volume to $-1$, to strongly contrast against the $1$'s in the volumes of the free parts.

During test, novel shapes can be generated from a random vector of all parts (Figure~\ref{fig:test}), where the anchor part can be randomly selected or user-specified.

\subsection{Training details}
We train and test our model on the ShapeNet part dataset~\cite{Yi16}, which is a subset of
the ShapeNet dataset~\cite{ShapeNet2015} and provides consistent alignment and semantic labeling for all shapes.
In the main paper, we test on four representative categories exhibiting rich part structure variation, including \emph{chairs} (3746), \emph{airplanes} (2690), \emph{lamps} (1546), \emph{motorbikes} (202).
\supl{The results on more categories is provided in the supplemental material.}
Each object category has a fixed number of semantic parts: a chair contains a \emph{back}, a \emph{seat}, a \emph{leg} and an \emph{armrest}; an airplane consists of a \emph{body}, a \emph{wing}, a \emph{tail} and an \emph{engine}; a lamp has a \emph{base}, a \emph{shade}, and a \emph{tube}; a motorbike is composed of a \emph{light}, a \emph{gas tank}, a \emph{seat}, a \emph{body}, a \emph{wheel} and a \emph{handle}. Note that a shape may not contain all semantic parts belonging to the corresponding category.
The dataset is divided into two parts, according to the official training/test split.
To enhance the training set, we employ the structure-aware deformation technique in~\cite{zheng2011component} to deform each shape, generating about $10$ variations of the shape. Finally, each shape and its semantic parts are voxelized to form our training set.

To train the part generators, we augment the dataset of part volumes via randomly scaling and translating the parts, with the ranges in $[0.5,1.5]$ for scaling and $[-20,20]$ (in voxels) for translation.
As Wasserstein GANs usually have large gradients, which might result in unstable training, we first pre-train the VAEs and then fine-tune them via joint training with the discriminators.
To train the part assembler, we generate a large set of training pairs of messed-up part arrangement and ground-truth assembly, with randomly selected anchor part. The messed-up arrangements are generated by randomly scaling and translating the semantic parts of a shape. The inverse of the messing-up transformations are used as ground-truth assembling transformations. Besides that, we also introduce some random noise to the training part volumes, to accommodate the imperfect volume generation during testing.

For all modules, we use ADAM ($\beta = 0.5$) for network optimization with an initial learning rate of $0.001$. Batch size is set to $32$. The parameters in the loss in Equation (\ref{eq:vae_loss}) are set as $\alpha_1=2$ and $\alpha_2=1\times10^{-3}$ for all experiments. The $\lambda$ in Equation (\ref{eq:gan_loss}) is set to $10$ as in~\cite{GulrajaniNIPS17}.
The average training time is $12$ hours for each part generator and $7$ hours for part assembler.

Note that the part generators and part assembler are not trained jointly, in an end-to-end fashion, since
there is no ground-truth assembly for the parts generated by VAE (random generation). It is, however, possible to make the whole pipeline end-to-end trainable if a discriminator network could be devised to judge whether the \emph{final assembled shape} is reasonable or not. We leave this for future work.


\section{Results and Evaluations}
\label{sec:result}

\vspace{-2pt}
\paragraph{Part generation.}
Symmetry preservation is an important aspect of structure reservation.
Through imposing reflective symmetry regularization for those parts which are reflectively symmetric, our model is able to produce structurally more plausible shapes.
To evaluate symmetry preservation in shape generation, we define a \emph{symmetry measure} for generated shapes. Given a generated shape volume, the reflective plane is the vertical bisector plane of the volume, since all training data were globally aligned and centered before voxelization. The symmetry measure can be obtained simply by reflecting the left half of the shape volume and computing the IoU against the right half.

Table~\ref{tab:ref_sym} shows the average symmetry measures on $1000$ randomly generated shapes.
The results are reported both for full shape and individual semantic parts.
We also compare to a baseline model trained without symmetry loss, as well as 3D-GAN~\cite{wu2016learning} and G2L~\cite{G2L18}. In G2L, an extra part refinement network is trained via minimizing the average reconstruction loss against three nearest neighbors retrieved from the training set. While achieving a higher symmetry score, such refinement also limited the diversity of the generated shapes, which is indicated by the lower inception score than ours in Table 3.

An interesting feature of our part generators is that it learns when to impose symmetry constraint on the generated parts, through judging from the input random vectors. This is due to the discriminative treatment of reflectively symmetric and asymmetric parts during the generator training (Equation (\ref{eq:vae_loss})). Take the leg part generator for example. If a random vector of a four-leg chair is input, the leg generator will preserve the reflective symmetry in the leg part. If the input random vector implies a swivel chair, the symmetry preservation will be automatically disabled since the leg part of a swivel chair is mostly \emph{not} reflectively symmetric in reality.
This is reflected in the average symmetry measures of symmetric and asymmetric legs in Table~\ref{tab:ref_sym}.

\begin{table}[t]
	\centering
	\small
	\caption{Comparing average symmetry measure over $1000$ generated shapes between our method
		(with and w/o symmetry loss) and 3D-GAN~\cite{wu2016learning}, G2L~\cite{G2L18}, on two shape categories. For each category, we report
		the measure for both full shape and semantic parts (\hlr{red} indicates the best and \hlg{green} the second best). Note how our method is able to discriminate between reflectively symmetric and asymmetric legs of chairs in preserving symmetries.}
	\label{tab:ref_sym}\vspace{-6pt}
	\scalebox{0.82}{
    \setlength{\tabcolsep}{1.0mm}{
		\begin{tabular}{l|l|c|c|c|c|c}
			\hline
			\multicolumn{2}{l|}{\multirow{2}{*}{}} & \multicolumn{2}{c|}{PAGENet} & \multirow{2}{*}{3D-GAN} & \multirow{2}{*}{G2L} & \multirow{2}{*}{GT}  \\
			\cline{3-4}
			\multicolumn{2}{l|}{}                  & w/ sym. loss & w/o sym. loss & & &\\
			\hline\hline
			\multirow{5}{*}{Chair} & back & $\hlg{0.88}$ & $0.80$ & $0.71$ & $\hlr{0.93}$ &$0.93$  \\
			\cline{2-7}
			& seat & $\hlg{0.90}$ & $0.82$ & $0.76$ & $\hlr{0.94}$ & $0.95$ \\
			\cline{2-7}
			& leg (sym.) & $\hlg{0.68}$ & $0.56$ & $0.40$ & $\hlr{0.74}$ &$0.86$\\
			\cline{2-7}
			& leg (asym.) & $0.29$ & $0.28$ & $-$ & $-$ & $0.32$ \\
			\cline{2-7}
			& armrest & $\hlr{0.64}$ & $\hlg{0.59}$ & $0.16$ & $\hlr{0.64}$ & $0.81$ \\
			\cline{2-7}
			& {\bf full} & $\hlg{0.85}$ & $0.77$ & $0.70$ & $\hlr{0.91}$ &$0.92$ \\
			\hline
			\multirow{4}{*}{Airplane} & body & $\hlg{0.84}$ & $0.78$ & $0.67$ & $\hlr{0.89}$ &$0.92$ \\
			\cline{2-7}
			& wing & $\hlg{0.81}$ & $0.73$ & $0.63$ &$\hlr{0.84}$ &$0.86$ \\
			\cline{2-7}
			& tail & $\hlg{0.75}$ & $0.69$ & $0.46$ & $\hlr{0.77}$ & $0.85$ \\
			\cline{2-7}
			& engine & $\hlr{0.71}$ & $\hlg{0.61}$ & $0.14$ & $\hlr{0.71}$ & $0.83$  \\
			\cline{2-7}
			& {\bf full} & $\hlg{0.80}$ & $0.72$ & $0.64$  &$\hlr{0.84}$ &$0.89$  \\
			\hline
		\end{tabular}
	}}\vspace{-6pt}
\end{table}

\begin{figure}[t]
	\begin{center}
		\includegraphics[width=1.0\linewidth]{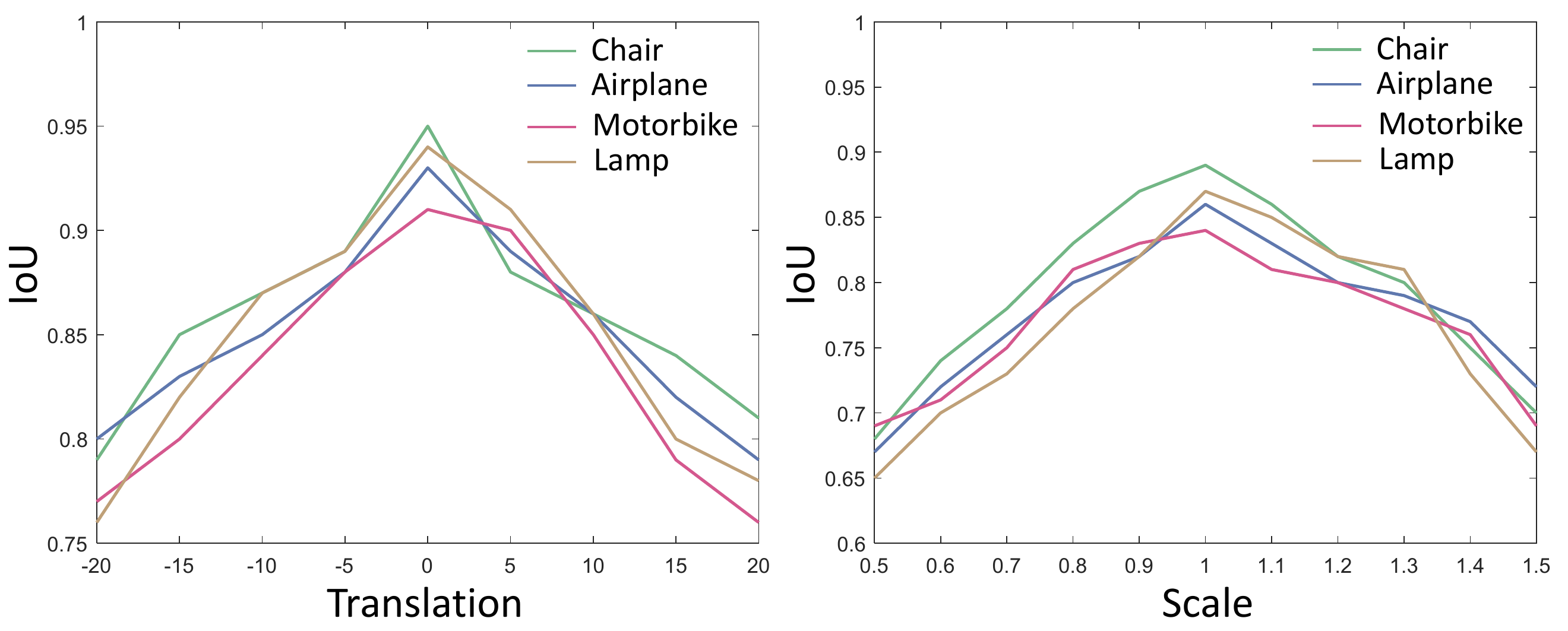}
	\end{center}
    \vspace{-12pt}
	\caption{Plots of assembly quality measure (average IoU w.r.t. ground-truth) over varying amount of translation (left; with scaling being fixed to $1.2$) and scaling (right; with translation being fixed to $10$).}
	\label{fig:assembly_plot}\vspace{-12pt}
\end{figure}

\begin{table*}[t!]
	\centering
	\small
	\caption{Comparison of part assembly quality (\hlr{red} indicates the best and \hlg{green} the second best).}
	\label{table:assembly_evalu}\vspace{-6pt}
	\scalebox{0.82}{
		\begin{tabular}{c |c| c| c| c|c| c| c| c| c|c|c|c|c|c|c|c|c}
			\hline
			&\multicolumn{4}{c|}{Chair} & \multicolumn{4}{|c|}{Plane} & \multicolumn{6}{|c|}{Motorbike}&\multicolumn{3}{|c}{Lamp}\\
			\hline\hline
			Template-based &\multicolumn{4}{c|}{0.74} & \multicolumn{4}{|c|}{0.72} & \multicolumn{6}{|c|}{0.69}&\multicolumn{3}{|c}{0.70}\\
			\hline\hline
			\multicolumn{1}{c|}{Anchor part} & seat & back & leg & armrest & body & wing& tail& engine& light& gas tank & seat & handle & wheel & body & base & shade & tube\\
			\hline		
			One-hot vector & 0.77 & 0.78 & 0.79 & 0.72 & 0.76 & 0.75 & 0.73 &0.74 & 0.72 & 0.76 & 0.74 & 0.73 & 0.71 & 0.76 & 0.79 & 0.77 & 0.70\\
			\hline
			Ours & \hlg{0.83} & \hlg{0.81} & \hlg{0.82} & \hlg{0.79} & \hlg{0.80} & \hlg{0.82} & \hlg{0.76} & \hlg{0.82} & \hlg{0.80} & \hlg{0.79} & \hlg{0.82} & \hlg{0.79} & \hlg{0.78} & \hlg{0.82} & \hlg{0.81} & \hlg{0.83} & \hlg{0.77}\\
			\hline
			Ours (training data) & \hlr{0.89} & \hlr{0.91} & \hlr{0.90} & \hlr{0.88} & \hlr{0.89} & \hlr{0.87} & \hlr{0.90} & \hlr{0.88} & \hlr{0.87} & \hlr{0.92} & \hlr{0.86} & \hlr{0.86} & \hlr{0.85} & \hlr{0.87} & \hlr{0.91} & \hlr{0.89} & \hlr{0.81}\\
			\hline
	\end{tabular}}\vspace{-6pt}
\end{table*}

\begin{figure*}[t]
	\begin{center}
		\includegraphics[width=0.95\linewidth]{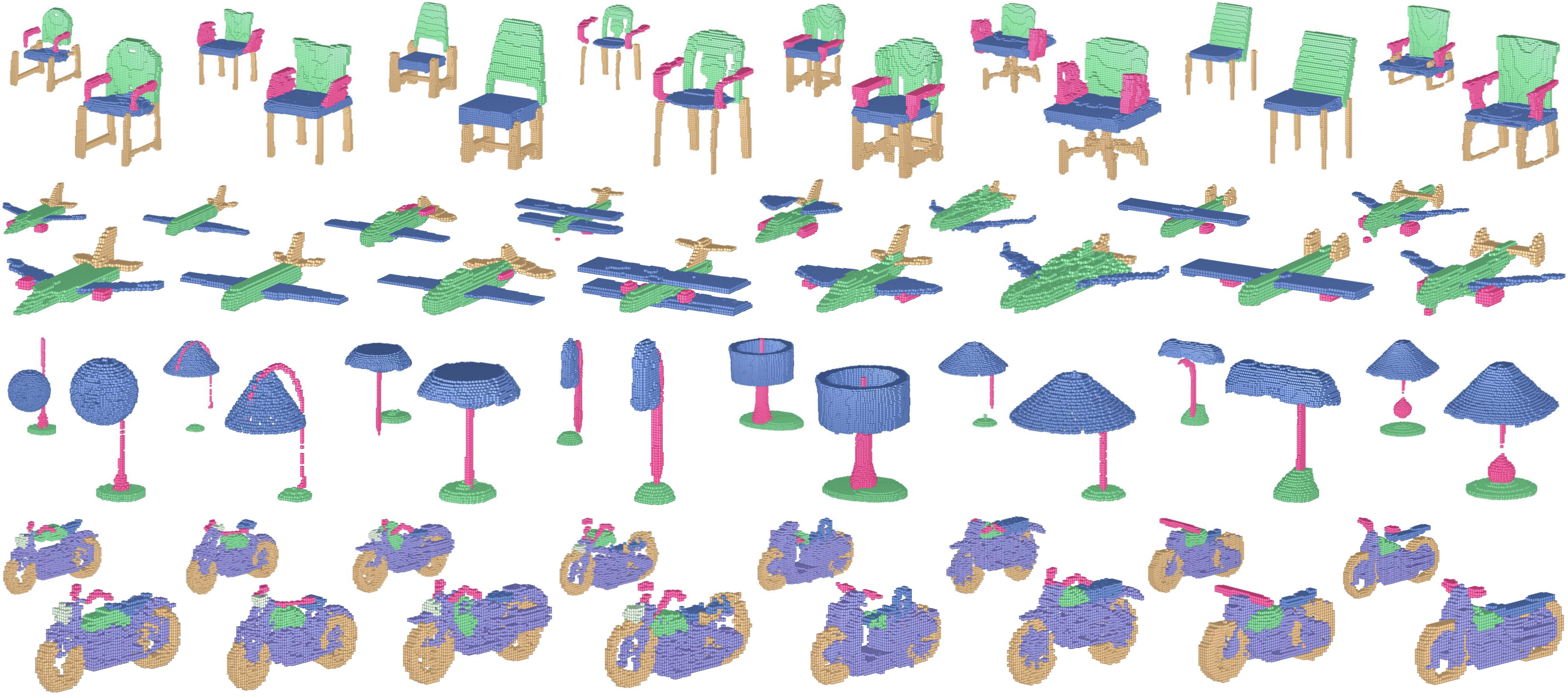}
	\end{center}\vspace{-10pt}
	\caption{Results of random shape generation on four categories. All generated shapes have semantic segmentation.}
	\label{fig:generation}\vspace{-8pt}
\end{figure*}

\vspace{-8pt}
\paragraph{Part assembly.}
To evaluate the ability of our part assembler, for each test shape, we perturb each of its semantic parts with random scaling and translation, and use our network to regress assembly transformation. Assembly quality is measured by the IoU between the assembled shape volume and the ground-truth. In testing, we choose each semantic part as anchor and report the average IoU as the assembly quality.
In Table~\ref{table:assembly_evalu}, we compare the assembly performance over \emph{three} methods.
The \emph{first} is our method. The \emph{second} is our method in which the anchor part is indicated by a one-hot vector. The \emph{third} one is a template-based part assembly where we retrieve a template shape from the training set based on part-wise CNN features. We then transform the shape parts according to the corresponding parts in the template, since part correspondence is available for all shapes in the training set and the generated shapes.
For contrasting, we also show the performance on the training shapes (the last row).

\begin{figure}[t]
	\begin{center}
		\includegraphics[width=\linewidth]{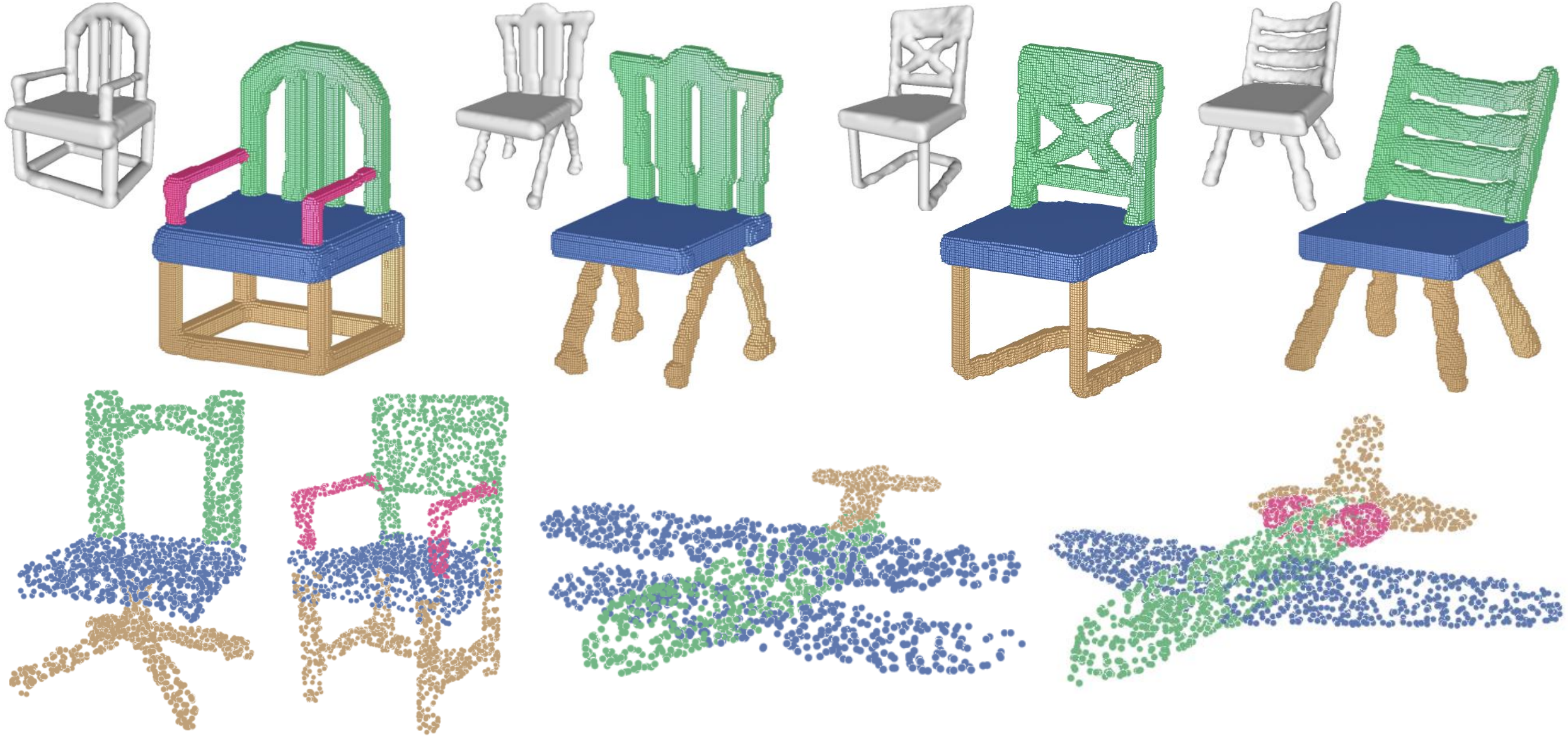}
	\end{center}
    \vspace{-8pt}
	\caption{Examples of high resolution ($128^3$) generation (top; the reconstructed surface mesh is shown for each shape) and point cloud generation (bottom).}
    \vspace{-12pt}
	\label{fig:hirespc}
\end{figure}

The results show that our part assembly generalizes well to unseen shapes. The numbers reported in Table~\ref{table:assembly_evalu} are under messy part arrangement with random scale from $[0.5,1.5]$ and random translation from $[-20,20]$.
For each part, its local axes are scaled and translated independently.
Figure~\ref{fig:assembly_plot} plots the assembly quality measure over varying amount of translation and scaling.
Our method obtains reasonably good assembly results within the range of $[-20,20]$ for translation and $[0.5,1.5]$ for scaling.
Note, however, the goal of our part assembler is \emph{not} to reconstruct an input shape. In fact, there is not a unique solution to structurally plausible part assembly. Therefore, this experiment only approximately evaluates the assembly ability of our model.

\begin{figure}[t]
	\begin{center}
		\includegraphics[width=\linewidth]{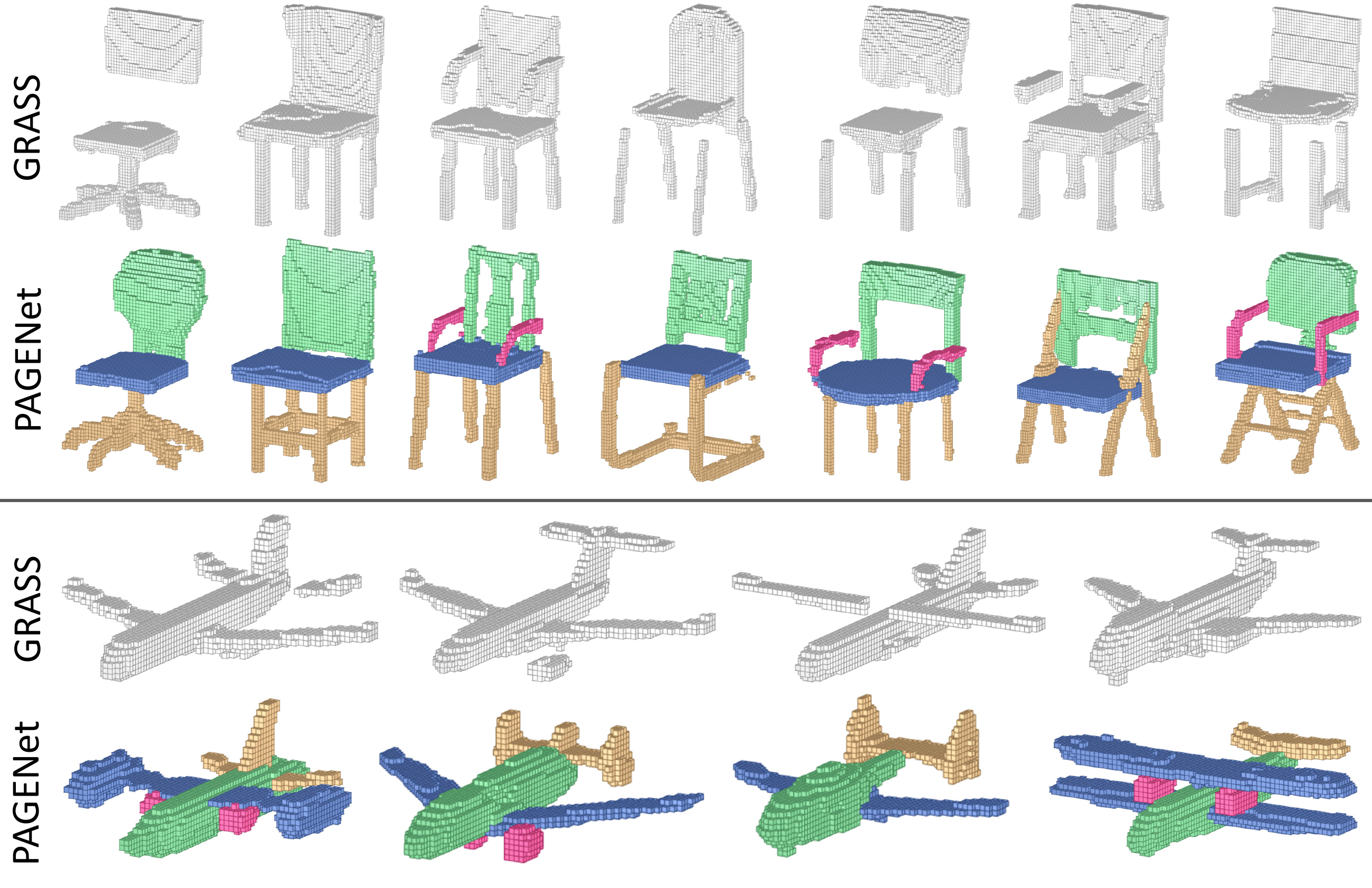}
	\end{center}
    \vspace{-8pt}
	\caption{A visual comparison of part connection with GRASS~\cite{li2017grass}.}
	\label{fig:comp_grass}
\end{figure}

\paragraph{Random shape generation.}
Figure~\ref{fig:generation} shows a few examples of random generation for all four shape categories.
For each shape, both the generate part volumes (overlaid) and the final assembling result are shown. A nice feature of our method is that the generated shapes all possess semantic segmentation by construction, which can be used in training data enhancement for shape segmentation. \supl{More generation results can be found in the supplemental material.}

In Table~\ref{tab:inception}, we compare the diversity of random generation by our method and three alternatives including 3D-GAN~\cite{wu2016learning}, GRASS~\cite{li2017grass} and G2L~\cite{G2L18}. Similar to~\cite{G2L18}, we use the inception score~\cite{salimans2016improved} to measure the diversity of shape sets.
In particular, we first cluster the training shapes and then train a classifier targeting the clusters.
The inception score for a given set of shapes is then measured based on the confidence and variance of the classification over the set. From the results, our method achieves consistently more diverse generation than alternatives, thanks to the part-wise shape variation modeling.
\supl{Please refer to the supplemental material for a \emph{qualitative} study on diversity.}

\begin{table}[t]\centering\small
\caption{Comparing diversity (inception score) of random shape generation with three state-of-the-art methods.}
\label{tab:inception}\vspace{-8pt}
\scalebox{1.0}{
\setlength{\tabcolsep}{1.2mm}{
\begin{tabular}{l|c|c|c|c}
\hline
                            & Chair     & Airplane  & Motorbike     & Lamp   \\ \hline\hline
3DGAN~\cite{wu2016learning} & $5.84$    & $5.61$    & $5.01$        & $5.19$  \\ \hline
GRASS~\cite{li2017grass}    & $5.79$    & $5.68$    & $4.93$        & $5.26$  \\ \hline
G2L~\cite{G2L18}            & $6.17$    & $5.89$    & $5.52$        & $5.45$  \\ \hline
PAGENet                     & $\bf{6.55}$    & $\bf{6.05}$    & $\bf{5.80}$        & $\bf{5.59}$  \\ \hline
\end{tabular}
}}\vspace{-12pt}
\end{table}

\vspace{-8pt}
\paragraph{High-res. and point cloud generation.}
The split of part synthesis and part assembly in our approach well supports high-resolution shape generation.
We first synthesize each part in very high resolution, within a local volume around the part. The synthesized
parts are then placed into a low-resolution global volume, in which the part assembler estimates an
assembling transformation for each part. The transformed parts are then unified in a high-resolution volume, resulting
in a high-res 3D model. Figure~\ref{fig:hirespc}(top) shows four examples of high-res shape generation. Through implementing the part generators with point cloud representation~\cite{Achlioptas2018},
PAGENet can also support 3D point cloud generation; see Figure~\ref{fig:hirespc}(bottom).

\begin{table}[b]\centering\small\vspace{-12pt}
\caption{Comparing with GRASS~\cite{li2017grass} on objective and subjective correctness rate of part assembly (in $\%$).}
\label{tab:comp_grass}\vspace{-6pt}
\setlength{\tabcolsep}{1.2mm}{
\begin{tabular}{l|c|c|c|c|c|c|c|c}
\hline
\multirow{2}{*}{} & \multicolumn{2}{c|}{Chair} & \multicolumn{2}{c|}{Airplane} & \multicolumn{2}{c|}{Motorbike} & \multicolumn{2}{c}{Lamp} \\ \cline{2-9}
                  & Obj.        & Subj.        & Obj.          & Subj.         & Obj.          & Subj.          & Obj.        & Subj.       \\ \hline\hline
GRASS             & $81.8$        & $79.6$         & $83.7$           & $80.2$          & $93.6$           & $90.1$            & $91.4$         & $87.4$        \\ \hline
PAGENet           & $90.2$         & $88.2$          & $92.6$           & $91.8$           & $97.0$           & $95.2$            & $96.8$         & $95.0$         \\ \hline
\end{tabular}}
\end{table}

\begin{figure}[t]
	\begin{center}
		\includegraphics[width=0.9\linewidth]{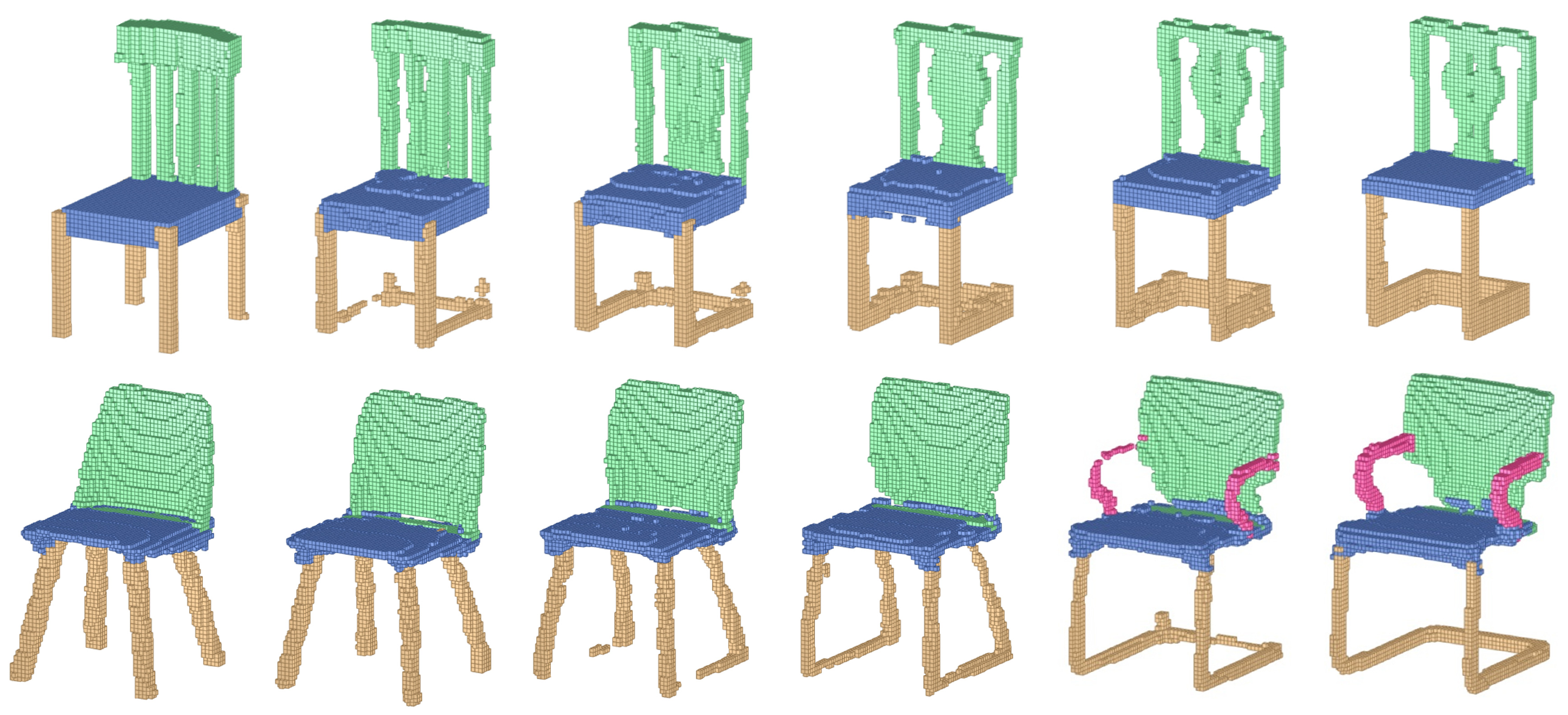}
	\end{center}
    \vspace{-12pt}
	\caption{Shape interpolation in latent space.}
	\label{fig:interpolation}
\end{figure}

\begin{figure}[t]
	\begin{center}
		\includegraphics[width=0.95\linewidth]{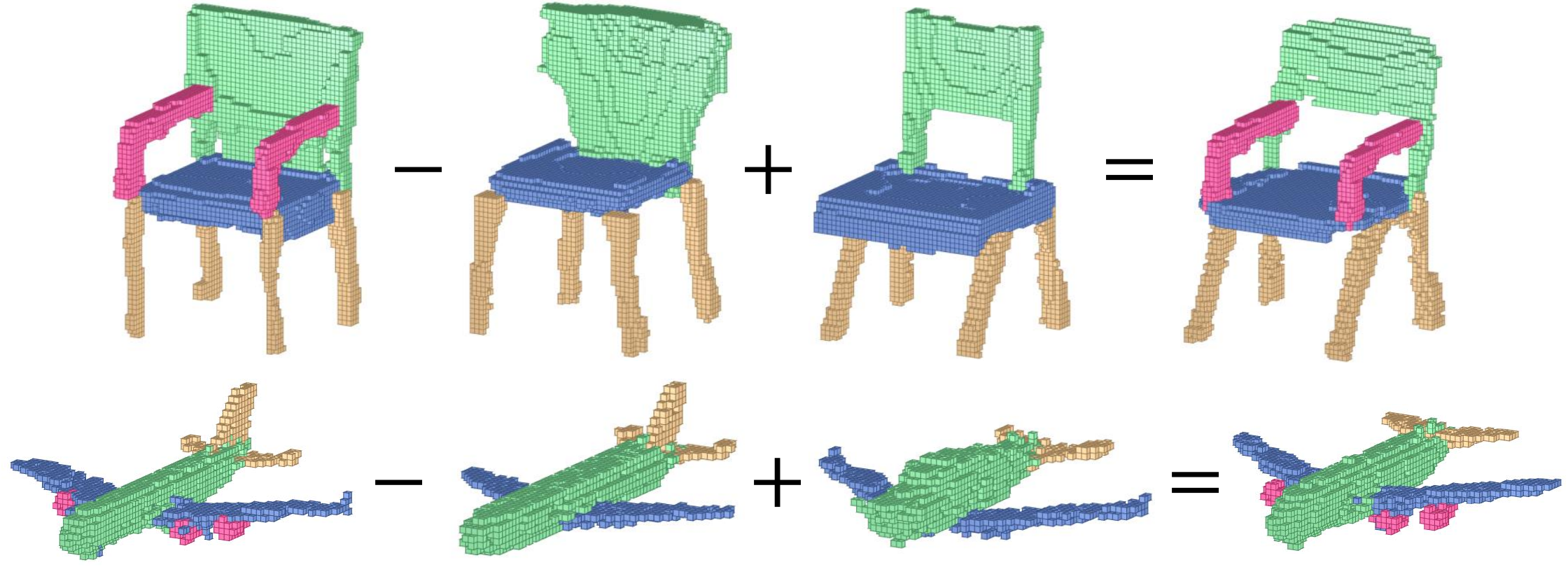}
	\end{center}
    \vspace{-12pt}
	\caption{Arithmetic operations in latent space.}
	\label{fig:arithmetic}\vspace{-12pt}
\end{figure}

\vspace{-8pt}
\paragraph{Comparison with GRASS~\cite{li2017grass}.}
Figure~\ref{fig:comp_grass} shows a visual comparison to GRASS.
Although GRASS can recover part relations (adjacency and symmetry) in the generated shapes, it does not explicitly
learn how adjacent parts are connected.
Therefore, parts generated by GRASS can sometimes mistakenly detach.
In contrast, PAGENet leads to better part connection thanks to the learned part assembly prior.
To quantitatively evaluate part assembly quality, we propose two measures, one objective and one subjective.
The objective measure simply examines the voxel connectivity of a generated shape volume.
In the subjective evaluation, we recruited $30$ human participants
(including both CS and non-CS graduate students) to
visually inspect and rate the correctness (binary choice) of part connections of a generated shape.
All generated shapes are presented to the human viewers with the same rendering style without part coloring.
For both measures, we compute the average \emph{correctness rate} of part assembly for each shape category.
Table~\ref{tab:comp_grass} compares the success rate of the two methods over $500$ randomly generated shapes for each category.

\vspace{-8pt}
\paragraph{Shape interpolation.}
Through breaking down 3D shape generation into part-wise generation and part assembly inference, our method is able to model significant variation of 3D shape structures. This can be demonstrated by shape interpolation between shapes with significantly different structures. Figure~\ref{fig:interpolation} shows two such examples. Our method can generate high quality in-between shapes with detailed shape structures, although the source and target shapes have considerably different structures.
\supl{More interpolation results can be found in the supplemental material.}

\vspace{-8pt}
\paragraph{Arithmetic in latent space.}
Due to the part-aware representation in our model, the latent space for full shapes is by construction structured. Therefore, our model naturally supports arithmetic modeling of 3D shapes, at the granularity of semantic part. Figure~\ref{fig:arithmetic} shows two examples arithmetic modeling. Again, the final shapes possess detailed shape structures due to part-aware generation. Meanwhile, the overall structures look plausible although the parts are originally from different shapes.

\begin{figure}[t]
	\begin{center}
		\includegraphics[width=0.9\linewidth]{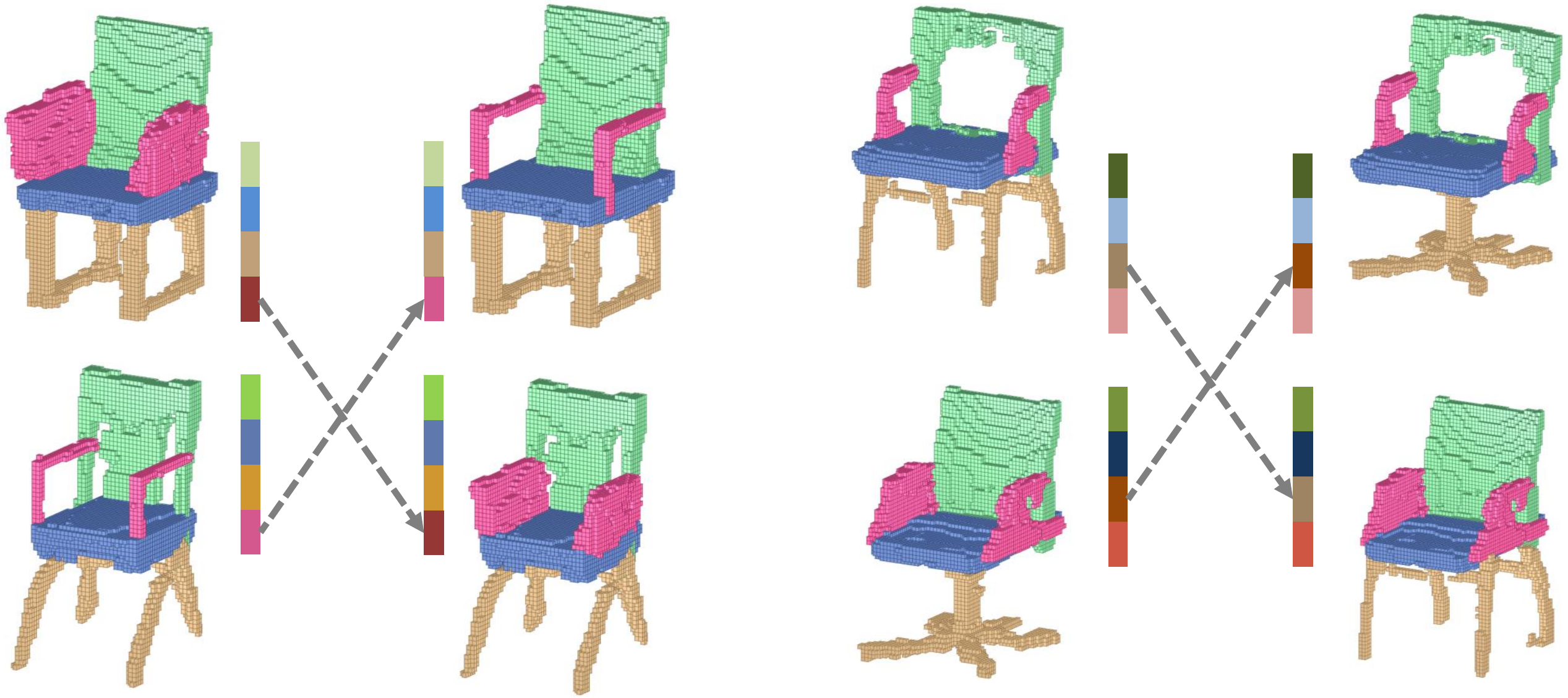}
	\end{center}\vspace{-12pt}
	\caption{3D shape crossover at the granularity of semantic part, enabled by our structured latent space.}
	\label{fig:crossover}\vspace{-8pt}
\end{figure}

\begin{figure}[t]
	\begin{center}
		\includegraphics[width=1.0\linewidth]{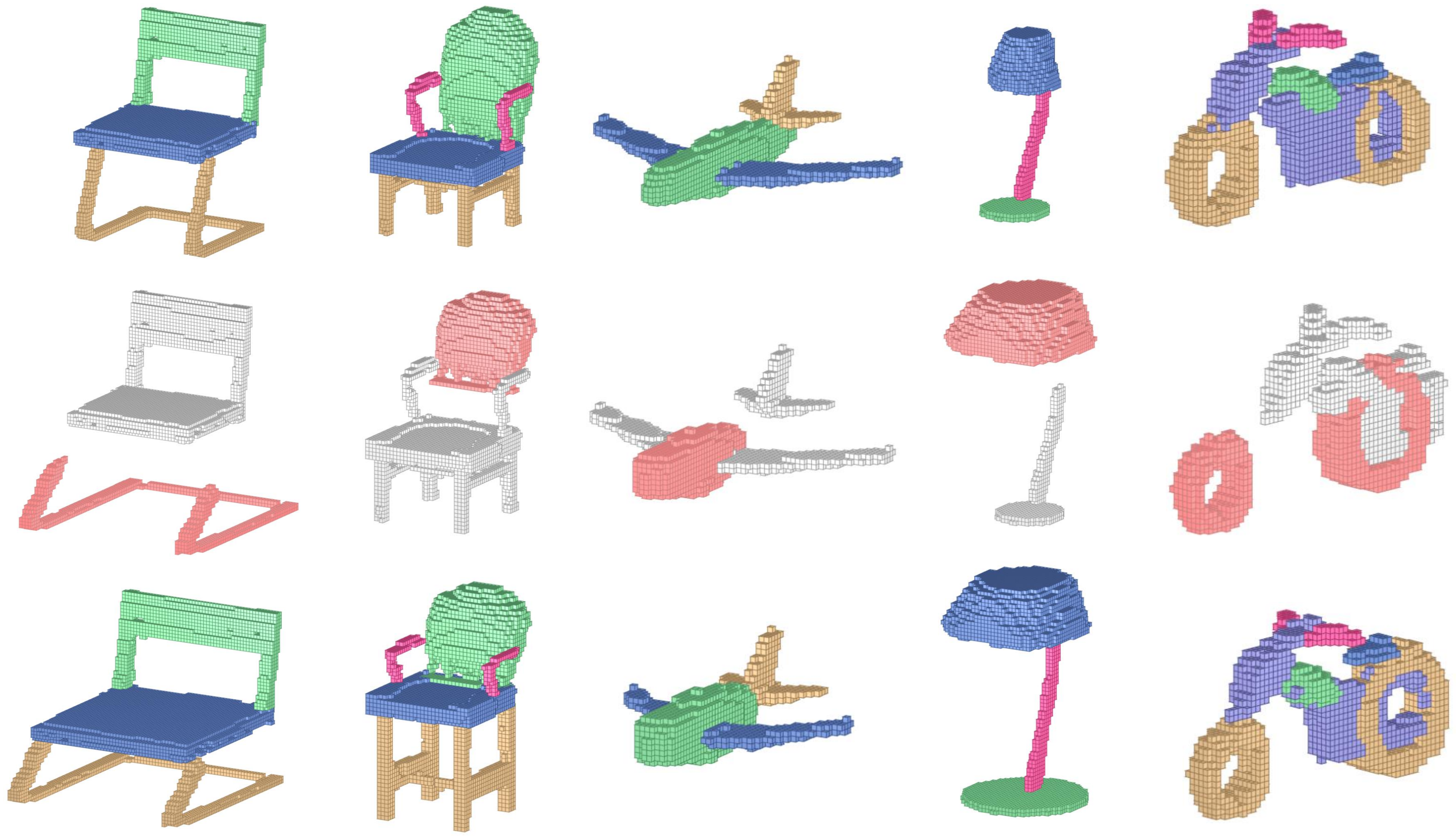}
	\end{center}\vspace{-12pt}
	\caption{Top: 3D shapes generated by PAGENet. Middle: User editing on anchor parts (pink). Bottom: Editing results after re-assembling.}
	\label{fig:shape_editing}\vspace{-12pt}
\end{figure}

\begin{figure}[t]
	\begin{center}
		\includegraphics[width=0.95\linewidth]{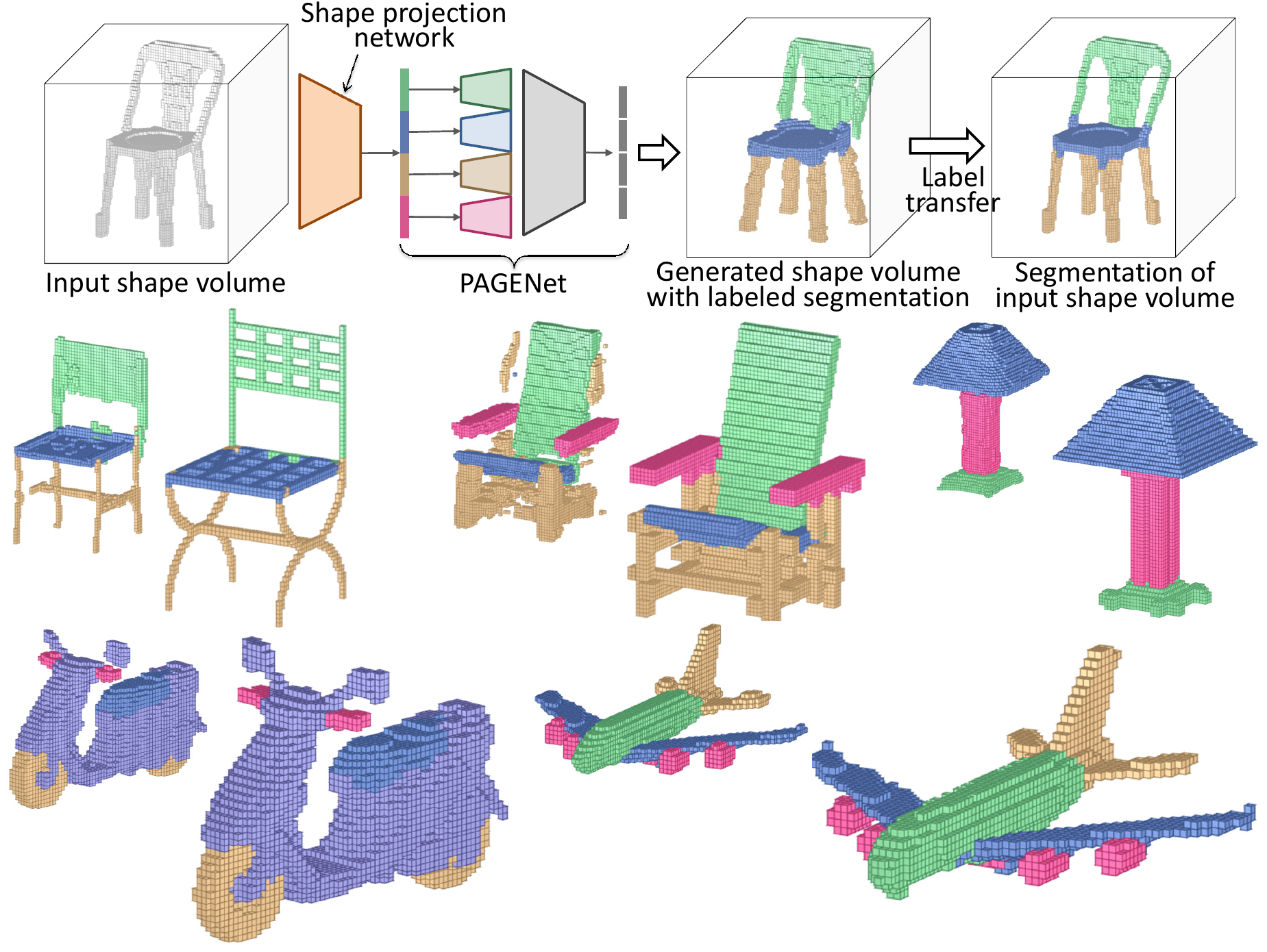}
	\end{center}\vspace{-12pt}
	\caption{Top: The network of using PAGENet for 3D shape segmentation. Bottom: A few segmentation results. For each example, the recovered shape used for semantic label transfer is shown to the top-left.}
	\label{fig:segmentation}\vspace{-8pt}
\end{figure}


\section{Applications}

\paragraph{Structure-aware part-based shape editing.}
This is an important way of creating 3D shape variations.
Our model supports two modes of part-based editing: \emph{latent space control} and \emph{direct manipulation}.
With the structured latent representation learned by our model, novel shapes can be created via
varying the entries of the latent vector of a generated shape.
This way, the latent code can be viewed as ``shape DNA'' of 3D shapes on which
chromosome operations such as mutation and crossover can be conducted.
New shapes are recovered (decoded) from the resultant chromosomes (code) in analogous to genetic expression,
facilitating bio-inspired shape generation~\cite{xu2012fit}.
Figure~\ref{fig:crossover} shows two examples of crossover operation.
%
Given a shape generated by our method, the \emph{direct manipulation} mode allows
the user to select one part as the anchor and perform direct editing (scaling and translation in our implementation) over it. After user editing, we leverage the part assembler to compute for each part an assembling
transformation, forming a structurally valid shape. Figure~\ref{fig:shape_editing} demonstrates a few examples of interactive part-based shape editing.

\vspace{-8pt}
\paragraph{Shape segmentation.}
The part-wise generation of our model can also be used to segment a 3D shape. The network is shown in Figure~\ref{fig:segmentation} (top).
Given a 3D shape in volumetric representation, we first project it into the latent space of PAGENet based on a trained shape projection network. It encodes the shape volume with five volumetric convolutional layers and project the input shape volume to a latent code. Then, our pre-trained PAGENet is used to reconstruct a 3D shape (with semantic segmentation). The projection network is trained by minimizing the reconstruction loss against the input shape volume, while keeping the PAGENet part fixed.
During testing, passing a 3D shape volume into the network results in a reconstructed 3D volume with semantic  segmentation. Since the recovered 3D volume is geometrically close to the input volume (due to the self-reconstruction training), we can accurately transfer its voxel labels onto the input volume, thus obtaining a semantic segmentation for the input.

\begin{table}[t!]
    \centering \small
    \caption{A quantitative comparison of shape segmentation.}
    \label{table:segmentation}\vspace{-6pt}
    \scalebox{0.96}{\setlength{\tabcolsep}{1.0mm}{
	\begin{tabular}{l|c|c|c|c}
		\hline
		& Chair & Airplane & Motorbike & Lamp  \\ \hline\hline
		PointNet~\cite{Su_CVPR17} & $89.6$ &$83.4$  & $65.2$ &$80.8$    \\ \hline
		SyncSpecCNN~\cite{yi2017syncspeccnn} &$90.2$  &$81.6$ & $\bf{66.7}$ & $84.7$ \\ \hline
		O-CNN~\cite{Wang-2017-OCNN} &$91.1$  &$\bf{85.5}$  & $56.9$ & $83.3$ \\ \hline
		PAGENet &$\bf{92.0}$ &$83.5$  & $63.2$& $\bf{85.2}$  \\ \hline
	\end{tabular}
    }}\vspace{-12pt}
\end{table}

Figure~\ref{fig:segmentation} shows a few examples of such segmentation on the testing set.
Essentially, we learn a deep model of \emph{segmentation transfer}. The model integrates a pre-learned \emph{part-aware shape manifold} with a \emph{shape-to-manifold projector}: To segment a shape, the projector retrieves the nearest neighbor from the manifold and generate semantic segmentation for the retrieved shape, whose segmentation can be easily transferred to the input due to shape resemblance. Table~\ref{table:segmentation} shows some quantitative results of segmentation with comparison with a few state-of-the-art methods. It shows that our method achieves comparable performance with the state-of-the-arts.
\supl{Extended results of shape segmentation can be found in the supplemental material.}


\section{Conclusion}
We have proposed a simple and effective generative model for quality and diverse 3D shape generation. Our model knows what it generates (semantic parts) and how the generated entities correlate with each other (by assembling transformation). This makes the generation part-aware and structure-revealing.
There are two main \emph{limitations}. \emph{First},
our model relies on hard-coded split of semantic part generators which is hard to adapt to a different label set.
\emph{Second}, our method currently works with major semantic parts. Extending it to deal with fine-grained parts would find difficulty in part assembly, which could be alleviated with the help of a hierarchical part organization as in~\cite{zhu_siga18,yu2018partnet}.


\section*{Acknowledgments}
We thank Hao Wang and Hui Huang for providing the utility codes (e.g., inception score).
This work was supported in part by NSFC (61572507, 61532003, 61622212).

{\small
\bibliographystyle{aaai}
\bibliography{pagenet}

\begin{thebibliography}{}

\bibitem[\protect\citeauthoryear{Achlioptas \bgroup et al\mbox.\egroup
  }{2018}]{Achlioptas2018}
Achlioptas, P.; Diamanti, O.; Mitliagkas, I.; and Guibas, L.
\newblock 2018.
\newblock Learning representations and generative models for 3d point clouds.
\newblock {\em arXiv preprint arXiv:1707.02392}.

\bibitem[\protect\citeauthoryear{Arjovsky, Chintala, and
  Bottou}{2017}]{arjovsky2017wasserstein}
Arjovsky, M.; Chintala, S.; and Bottou, L.
\newblock 2017.
\newblock Wasserstein gan.
\newblock {\em arXiv preprint arXiv:1701.07875}.

\bibitem[\protect\citeauthoryear{Balashova \bgroup et al\mbox.\egroup
  }{2018}]{Balashova2018}
Balashova, E.; Singh, V.; Wang, J.; Teixeira, B.; Chen, T.; and Funkhouser, T.
\newblock 2018.
\newblock Structure-aware shape synthesis.
\newblock In {\em 3D Vision (3DV)}.

\bibitem[\protect\citeauthoryear{Chang \bgroup et al\mbox.\egroup
  }{2015}]{ShapeNet2015}
Chang, A.~X.; Funkhouser, T.; J.~Guibas, L.; Hanrahan, P.; Huang, Q.; Li, Z.;
  Savarese, S.; Savva, M.; Song, S.; Su, H.; Xiao, J.; Yi, L.; and Yu, F.
\newblock 2015.
\newblock {ShapeNet: An Information-Rich 3D Model Repository}.
\newblock (arXiv:1512.03012 [cs.GR]).

\bibitem[\protect\citeauthoryear{Chen and Zhang}{2019}]{chen2018learning}
Chen, Z., and Zhang, H.
\newblock 2019.
\newblock Learning implicit fields for generative shape modeling.
\newblock In {\em CVPR}.

\bibitem[\protect\citeauthoryear{Dubrovina \bgroup et al\mbox.\egroup
  }{2019}]{Dubrovina2019}
Dubrovina, A.; Xia, F.; Achlioptas, P.; Shalah, M.; and Guibas, L.
\newblock 2019.
\newblock Composite shape modeling via latent space factorization.
\newblock {\em arXiv preprint arXiv:1803.10932}.

\bibitem[\protect\citeauthoryear{Fan, Su, and Guibas}{2016}]{fan2016point}
Fan, H.; Su, H.; and Guibas, L.
\newblock 2016.
\newblock A point set generation network for 3d object reconstruction from a
  single image.
\newblock {\em arXiv preprint arXiv:1612.00603}.

\bibitem[\protect\citeauthoryear{Girdhar \bgroup et al\mbox.\egroup
  }{2016}]{girdhar2016learning}
Girdhar, R.; Fouhey, D.~F.; Rodriguez, M.; and Gupta, A.
\newblock 2016.
\newblock Learning a predictable and generative vector representation for
  objects.
\newblock In {\em European Conference on Computer Vision},  484--499.
\newblock Springer.

\bibitem[\protect\citeauthoryear{Groueix \bgroup et al\mbox.\egroup
  }{2018}]{groueix2018papier}
Groueix, T.; Fisher, M.; Kim, V.~G.; Russell, B.~C.; and Aubry, M.
\newblock 2018.
\newblock A papier-m{\^a}ch{\'e} approach to learning 3d surface generation.
\newblock In {\em CVPR},  216--224.

\bibitem[\protect\citeauthoryear{Gulrajani \bgroup et al\mbox.\egroup
  }{2017}]{GulrajaniNIPS17}
Gulrajani, I.; Ahmed, F.; Arjovsky, M.; Dumoulin, V.; and Courville, A.~C.
\newblock 2017.
\newblock Improved training of wasserstein gans.
\newblock In {\em NIPS},  5769--5779.

\bibitem[\protect\citeauthoryear{Huang, Kalogerakis, and
  Marlin}{2015}]{Huang:2015:deeplearningsurfaces}
Huang, H.; Kalogerakis, E.; and Marlin, B.
\newblock 2015.
\newblock Analysis and synthesis of 3d shape families via deep-learned
  generative models of surfaces.
\newblock {\em Computer Graphics Forum} 34(5).

\bibitem[\protect\citeauthoryear{Li \bgroup et al\mbox.\egroup
  }{2017}]{li2017grass}
Li, J.; Xu, K.; Chaudhuri, S.; Yumer, E.; Zhang, H.; and Guibas, L.
\newblock 2017.
\newblock Grass: Generative recursive autoencoders for shape structures.
\newblock {\em arXiv preprint arXiv:1705.02090}.

\bibitem[\protect\citeauthoryear{Mitra \bgroup et al\mbox.\egroup
  }{2013}]{mitra2013structure}
Mitra, N.; Wand, M.; Zhang, H.~R.; Cohen-Or, D.; Kim, V.; and Huang, Q.-X.
\newblock 2013.
\newblock Structure-aware shape processing.
\newblock In {\em SIGGRAPH Asia 2013 Courses}, ~1.
\newblock ACM.

\bibitem[\protect\citeauthoryear{Mitra, Guibas, and
  Pauly}{2006}]{mitra2006partial}
Mitra, N.~J.; Guibas, L.~J.; and Pauly, M.
\newblock 2006.
\newblock Partial and approximate symmetry detection for 3d geometry.
\newblock {\em ACM Trans. on Graph.} 25(3):560--568.

\bibitem[\protect\citeauthoryear{Mo \bgroup et al\mbox.\egroup
  }{2019}]{mo2019structurenet}
Mo, K.; Guerrero, P.; Yi, L.; Su, H.; Wonka, P.; Mitra, N.; and Guibas, L.~J.
\newblock 2019.
\newblock Structurenet: Hierarchical graph networks for 3d shape generation.
\newblock {\em ACM Trans. on Graph. (SIGGRAPH Asia)}.

\bibitem[\protect\citeauthoryear{Nash and Williams}{2017}]{Nash2017}
Nash, C., and Williams, C.~K.
\newblock 2017.
\newblock The shape variational autoencoder: A deep generative model of
  part-segmented 3d objects.
\newblock {\em Computer Graphics Forum (SGP 2017)} 36(5):1--12.

\bibitem[\protect\citeauthoryear{Park \bgroup et al\mbox.\egroup
  }{2019}]{park2019deepsdf}
Park, J.~J.; Florence, P.; Straub, J.; Newcombe, R.; and Lovegrove, S.
\newblock 2019.
\newblock Deepsdf: Learning continuous signed distance functions for shape
  representation.
\newblock In {\em CVPR}.

\bibitem[\protect\citeauthoryear{Riegler, Ulusoy, and
  Geiger}{2017}]{riegler2017octnet}
Riegler, G.; Ulusoy, A.~O.; and Geiger, A.
\newblock 2017.
\newblock Octnet: Learning deep 3d representations at high resolutions.
\newblock In {\em Proc. CVPR}, volume~3.

\bibitem[\protect\citeauthoryear{Salimans \bgroup et al\mbox.\egroup
  }{2016}]{salimans2016improved}
Salimans, T.; Goodfellow, I.; Zaremba, W.; Cheung, V.; Radford, A.; and Chen,
  X.
\newblock 2016.
\newblock Improved techniques for training gans.
\newblock In {\em Proc. NIPS},  2234--2242.

\bibitem[\protect\citeauthoryear{Schor \bgroup et al\mbox.\egroup
  }{2019}]{Schor2019}
Schor, N.; Katzier, O.; Zhang, H.; and Cohen-Or, D.
\newblock 2019.
\newblock Learning to generate the "unseen" via part synthesis and composition.
\newblock In {\em IEEE International Conference on Computer Vision}.

\bibitem[\protect\citeauthoryear{Soltani \bgroup et al\mbox.\egroup
  }{2017}]{ArsalanCVPR2017}
Soltani, A.~A.; Huang, H.; Wu, J.; Kulkarni, T.~D.; and Tenenbaum, J.~B.
\newblock 2017.
\newblock Synthesizing 3d shapes via modeling multi-view depth maps and
  silhouettes with deep generative networks.
\newblock In {\em Proceedings of the IEEE Conference on Computer Vision and
  Pattern Recognition},  1511--1519.

\bibitem[\protect\citeauthoryear{Su \bgroup et al\mbox.\egroup
  }{2017}]{Su_CVPR17}
Su, H.; Qi, C.~R.; Mo, K.; and Guibas, L.~J.
\newblock 2017.
\newblock Pointnet: Deep learning on point sets for 3d classification and
  segmentation.
\newblock In {\em Proc. IEEE Conference on Computer Vision and Pattern
  Recognition (CVPR)},  to appear.

\bibitem[\protect\citeauthoryear{Wang \bgroup et al\mbox.\egroup
  }{2017}]{Wang-2017-OCNN}
Wang, P.-S.; Liu, Y.; Guo, Y.-X.; Sun, C.-Y.; and Tong, X.
\newblock 2017.
\newblock {O-CNN: Octree-based Convolutional Neural Networks for 3D Shape
  Analysis}.
\newblock {\em ACM Transactions on Graphics (SIGGRAPH)} 36(4).

\bibitem[\protect\citeauthoryear{Wang \bgroup et al\mbox.\egroup
  }{2018a}]{G2L18}
Wang, H.; Schor, N.; Hu, R.; Huang, H.; Cohen-Or, D.; and Huang, H.
\newblock 2018a.
\newblock Global-to-local generative model for 3d shapes.
\newblock {\em ACM Transactions on Graphics (Proc. SIGGRAPH ASIA)}
  37(6):214:1—214:10.

\bibitem[\protect\citeauthoryear{Wang \bgroup et al\mbox.\egroup
  }{2018b}]{wang2018pixel2mesh}
Wang, N.; Zhang, Y.; Li, Z.; Fu, Y.; Liu, W.; and Jiang, Y.-G.
\newblock 2018b.
\newblock Pixel2mesh: Generating 3d mesh models from single rgb images.
\newblock In {\em ECCV},  52--67.

\bibitem[\protect\citeauthoryear{Wu \bgroup et al\mbox.\egroup
  }{2016}]{wu2016learning}
Wu, J.; Zhang, C.; Xue, T.; Freeman, B.; and Tenenbaum, J.
\newblock 2016.
\newblock Learning a probabilistic latent space of object shapes via 3d
  generative-adversarial modeling.
\newblock In {\em Advances in Neural Information Processing Systems},  82--90.

\bibitem[\protect\citeauthoryear{Wu \bgroup et al\mbox.\egroup
  }{2018}]{Wu2018Struct}
Wu, Z.; Wang, X.; Lin, D.; Lischinski, D.; Cohen-Or, D.; and Huang, H.
\newblock 2018.
\newblock Structure-aware generative network for 3d-shape modeling.
\newblock {\em arXiv preprint arXiv:1808.03981}.

\bibitem[\protect\citeauthoryear{Xu \bgroup et al\mbox.\egroup
  }{2012}]{xu2012fit}
Xu, K.; Zhang, H.; Cohen-Or, D.; and Chen, B.
\newblock 2012.
\newblock Fit and diverse: set evolution for inspiring 3d shape galleries.
\newblock {\em ACM Transactions on Graphics (TOG)} 31(4):57.

\bibitem[\protect\citeauthoryear{Xu \bgroup et al\mbox.\egroup
  }{2016}]{xu2016data}
Xu, K.; Kim, V.~G.; Huang, Q.; Mitra, N.; and Kalogerakis, E.
\newblock 2016.
\newblock Data-driven shape analysis and processing.
\newblock In {\em SIGGRAPH ASIA 2016 Courses}, ~4.
\newblock ACM.

\bibitem[\protect\citeauthoryear{Yi \bgroup et al\mbox.\egroup }{2016}]{Yi16}
Yi, L.; Kim, V.~G.; Ceylan, D.; Shen, I.-C.; Yan, M.; Su, H.; Lu, C.; Huang,
  Q.; Sheffer, A.; and Guibas, L.
\newblock 2016.
\newblock A scalable active framework for region annotation in 3d shape
  collections.
\newblock {\em SIGGRAPH Asia}.

\bibitem[\protect\citeauthoryear{Yi \bgroup et al\mbox.\egroup
  }{2017}]{yi2017syncspeccnn}
Yi, L.; Su, H.; Guo, X.; and Guibas, L.~J.
\newblock 2017.
\newblock Syncspeccnn: Synchronized spectral cnn for 3d shape segmentation.
\newblock In {\em CVPR},  6584--6592.

\bibitem[\protect\citeauthoryear{Yu \bgroup et al\mbox.\egroup
  }{2018}]{yu2018partnet}
Yu, F.; Liu, K.; Zhang, Y.; Zhu, C.; and Xu, K.
\newblock 2018.
\newblock Partnet: A recursive part decomposition network for fine-grained and
  hierarchical shape segmentation.
\newblock In {\em CVPR}.

\bibitem[\protect\citeauthoryear{Zheng \bgroup et al\mbox.\egroup
  }{2011}]{zheng2011component}
Zheng, Y.; Fu, H.; Cohen-Or, D.; Au, O. K.-C.; and Tai, C.-L.
\newblock 2011.
\newblock Component-wise controllers for structure-preserving shape
  manipulation.
\newblock In {\em Computer Graphics Forum}, volume~30,  563--572.
\newblock Wiley Online Library.

\bibitem[\protect\citeauthoryear{Zhu \bgroup et al\mbox.\egroup
  }{2018}]{zhu_siga18}
Zhu, C.; Xu, K.; Chaudhuri, S.; Yi, R.; and Zhang, H.
\newblock 2018.
\newblock Scores: Shape composition with recursive substructure priors.
\newblock {\em ACM Transactions on Graphics (SIGGRAPH Asia 2018)} 37(6):to
  appear.

\bibitem[\protect\citeauthoryear{Zou \bgroup et al\mbox.\egroup
  }{2017}]{zou20173d}
Zou, C.; Yumer, E.; Yang, J.; Ceylan, D.; and Hoiem, D.
\newblock 2017.
\newblock 3d-prnn: Generating shape primitives with recurrent neural networks.
\newblock In {\em The IEEE International Conference on Computer Vision (ICCV)}.

\end{thebibliography}
}

\end{document}